\documentclass[twoside,11pt]{article}

\usepackage{amsmath}
\usepackage{comment}
\newcommand{\indep}{\perp \!\!\! \perp}
\usepackage{blindtext}
\usepackage{amsmath}
\usepackage{mathtools}
\usepackage{amssymb,amsmath,amsthm}
\usepackage{amsmath}
\usepackage{algorithm}
\usepackage{algorithmic}

\usepackage{mathtools}
\usepackage{adjustbox}
\usepackage{array}
\usepackage{booktabs}
\usepackage{dsfont}
\usepackage{multirow}
\usepackage{hhline}
\usepackage[bottom]{footmisc}
\DeclareMathOperator*{\argmin}{arg\,min}
\DeclareMathOperator*{\argmax}{arg\,max}
\usepackage[abbrvbib, preprint]{jmlr2e}

\ShortHeadings{}{}
\firstpageno{1}
\usepackage{float}
\usepackage[caption = false]{subfig}
\usepackage{graphicx}

\begin{document}

\title{\emph{OptiGrad}: A Fair and more Efficient Price Elasticity Optimization via a Gradient Based Learning}

\author{\name Vincent Grari \email vincent.grari@axa.com \\
       \addr AXA Group Operations\\
       TRAIL, LIP6, Sorbonne Université, Paris, France \\
       \AND
        \name Marcin Detyniecki \email marcin.detyniecki@axa.com \\
       \addr AXA Group Operations\\
       \addr Polish Academy of Science, IBS PAN, Warsaw, Poland \\
       TRAIL, LIP6, Sorbonne Université, Paris, France \\}

\editor{My editor}

\maketitle
\begin{abstract}


This paper presents a novel approach to optimizing profit margins in non-life insurance markets through a gradient descent-based method, targeting three key objectives: 1) maximizing profit margins, 2) ensuring conversion rates, and 3) enforcing fairness criteria such as demographic parity (DP). Traditional pricing optimization, which heavily lean on linear and semi definite programming, encounter challenges in balancing profitability and fairness. These challenges become especially pronounced in situations that necessitate continuous rate adjustments and the incorporation of fairness criteria. Specifically, indirect Ratebook optimization, a widely-used method for new business price setting, relies on predictor models such as XGBoost or GLMs/GAMs to estimate on downstream individually optimized prices. However, this strategy is prone to sequential errors and struggles to effectively manage optimizations for continuous rate scenarios. In practice, to save time actuaries frequently opt for optimization within discrete intervals (e.g., range of [-20\%, +20\%] with 
fix increments) leading to approximate estimations. Moreover, to circumvent infeasible solutions they often use relaxed constraints leading to suboptimal pricing strategies. 
The reverse-engineered nature of traditional models complicates the enforcement of fairness 
and can lead to biased outcomes. Our method addresses these challenges by employing a direct optimization strategy in the continuous space of rates and by embedding fairness through an adversarial predictor model. This innovation not only reduces sequential errors and simplifies the complexities found in traditional models but also directly integrates fairness measures into the commercial premium calculation. 
We demonstrate improved margin performance and stronger enforcement of fairness highlighting  the critical need to evolve existing pricing strategies.

\end{abstract}


\section{Introduction}
The non-life insurance sector is currently experiencing a significant transition, driven by rapid evolution attributed to intensified competition
among insurers, regulations, and new technologies. These dynamics necessitate the agile development of new pricing strategies. Such new strategies must not only ensure profitability and foster customer acquisition and retention but also align  with an ethical paradigm that emphasizes fairness and equity, as underscored by recent legislative initiatives~\citep{EuropeanParliament2016a,com2021laying,ito2017optimization}. 
The insurance industry has long been anchored by traditional pricing methodologies such as individual and Ratebook optimizations, crucial for determining both new business commercial and renewals premiums. These methodologies rely heavily on linear and semi definite programming techniques, forming the analytical backbone of pricing strategies~\citep{deLarrard,verschuren2022customer,hashorva2018some}. These traditional approaches face increasing challenges in reconciling the threefold goals of profitability, customer retention/acquisition, and ethical considerations.

In real world operations, a common limitation of current pricing strategies is their dependence on discrete optimized rates  rather than exploring the full potential of continuous rate variations. This conventional approach is limited to a predefined set of options, narrowing the optimization scope, inevitably resulting in approximate estimations. 
A significant reason for this limitation is the extensive time necessary for an exhaustive exploration of continuous optimized rate. Furthermore, to address the challenges of infeasibility solutions within this constrained optimization framework, constraints are frequently relaxed~\citep{deLarrard,verschuren2022customer}. This can inadvertently lead to an increased risk of errors accumulating in the pricing process.  

On another dimension, the reliance on reverse-engineered predictive models in conventional Indirect Ratebook optimization complicates the enforcement of fairness. Traditionally, the prevailing fairness approach within the insurance pricing~\citep{lindholm2022discrimination,grari2022fair,xin2023antidiscrimination,lindholm2022discussion,lindholm2023fair} has emphasized ensuring fairness at the pure premium calculation stage only, where predictive models are designed to prevent claim frequency and cost predictions from being influenced by sensitive variables such as gender or race. 
However, we claim that the fairness enforcement is not guarantee at the commercial premium layer. In fact, by adding a second layer -  commercial premium - could compromise or negate the fairness achieved at the pure premium level. This occurs because the commercial layer integrates an additional objective, balancing profit margins with conversion rates, which can inadvertently reintroduce biases. 
For instance, in our experiment, we observed that increasing conversion rates increases biases against certain sensitive variables.
This new challenge requires new methods that enforces fairness directly at the final pricing layer.

This paper introduces \emph{OptiGrad}, a novel approach that employs a gradient descent learning strategy specifically tailored for profit margin optimization in the non-life insurance sector. \emph{OptiGrad} is designed with the explicit intent to achieve three core objectives: maximizing profit margins, ensuring a minimum conversion rate, and most importantly, enforcing fairness through criteria such as demographic parity (DP). This framework employs an offline \textbf{differentiable} conversion model and a \textbf{differentiable} 
pure premium model. 
The key intuition of \emph{OptiGrad} is  adopting neural network architecture principles, particularly the application of gradient descent, where the output of one model is optimized in conjunction with the input of other models. The optimization procedure leverages the chain rule to derive the optimal pricing rate by integrating inputs from both the conversion and pure premium models into the objective function, utilizing a differentiable model for estimation. 
The differentiability of these models, 
plays a crucial role in incorporating traditional fairness mechanisms. This particularity allows the use of techniques such as fair adversarial networks, which rely on gradient-based optimization, to be directly integrated into the objective function. Through experimental testing, \emph{OptiGrad} proves to enhance the Global Written Margin (GWM) while keeping the same conversion rate by simultaneously improving fairness at the market premium level, shifting the focus away from the traditional approach of targeting the pure premium. 

\section{Problem Statement}


Throughout this document, we consider $X$ as the set customer features, a variable $h_{w_h}(X)$ as the estimated 
actuarial risk. This estimation may represent frequency or average cost models, or the pure premium derived from the product of these two sub-models, all trained using a differentiable supervised machine learning algorithm 
$h_{w_h}$ with parameters $w_h$.
Furthermore, we introduce $f_{w_f}$ a differentiable conversion model that has been trained on a historical dataset X containing the actual conversion binary information $Y$. 
Among the features, $S$ will denote a sensitive attribute (e.g., gender or race feature), that cannot be used at test-time, but, 
is a variable that we must observe to ensure fairness of the model. 
Depending on the context, the sensitive attribute domain $\Omega_{S}$ of the sensitive attribute $S$ may be discrete or continuous. The training data consists of $n$ examples ${(x_{i},s_{i},y_{i})}$ 
that is sampled from a training distribution $p$, where $x_{i} \in \mathbb{R}^{d}$ is the feature vector with $d$-dimensional feature vector for the $i$-th policyholder, $s_i \in \Omega_{S}$ denotes the sensitive attribute's value, 
and $y_{i} \in \Omega_{Y}$ is the binary conversion label. 


\subsection{Price Optimization Methods}\label{sec:fair}

In optimizing commercial pricing, several strategies are deployed with distinctive methodologies. Individual Optimization, for instance, calculates the optimal price for each customer based on their unique risk profile. Alternatively, Indirect Ratebook Optimization applies reverse-engineering on downstream optimized prices, which are frequently the outcomes of Individual Optimization.
Each method has its own advantages and limitations, which can vary depending on whether the optimization is conducted in an online or offline environment. Individual Optimization, for example, is noted for its precision, eliminating the potential for compounded errors that may arise from the reverse-engineering process inherent in Indirect Ratebook Optimization. However, its practical application can be complex, with challenges in operationalizing rating prices due to the necessity of direct price calculations. On the other hand, Indirect Ratebook Optimization facilitates easier implementation within the insurance rating system for production, though its offline nature limits market responsiveness.

This study primarily focuses on Direct Ratebook Optimization, which seeks to combine the production deployment advantages of Indirect Optimization with a more straightforward method of price optimization, eliminating the need for reverse engineering. This method strives for an optimal balance between accuracy and operational feasibility. 

\subsubsection*{Individual optimization}
\label{sec:indopt}
Individual pricing optimization under constraints allows to maintain 
a consistent pricing strategy. 
Constraints may vary significantly by market, in some countries, regulatory frameworks enforce limits on price adjustments to prevent extreme  high prices, with specific thresholds set limits. 
Furthermore, it's crucial to consider the impact on vulnerable segments 
to avoid large price increases that could lead to customer dissatisfaction and negatively affect the brand. Therefore, optimization strategies need to navigate between maintaining equity and leveraging market dynamics, ensuring that price adjustments are justifiable and align with both business objectives and customer expectations.

The optimization program can then be described as below:

\begin{equation}
\begin{aligned}
& \argmax_{c_1,..,c_n}
& & \sum_{i=1}^{n}[(c_i * h_{w_h}(x_i) - h_{w_h}(x_i)) * f_{w_f}(x_i, c_i * h_{w_h}(x_i))] \\
& \text{s.t.} 
& & c_i  \leq b, \quad i = 1, \ldots, n. 
\\
& 
& & c_i   \geq a, \quad i = 1, \ldots, n. 
\\
& 
& & \frac{1}{n} \sum_{i=1}^{n} f_{w_f}(x_i, c_i * h_{w_h}(x_i)) \geq \gamma 
\label{eq:indopt}
\end{aligned}
\end{equation}

The commercial price $p_i = c_i * h_{w_h}(x_i)$ is defined via a commercial coefficient $c_i$ for each individual customer $i$. The predictor model $h_{w_h}$ takes as input the customers features $X$. The conversion probability $f_{w_f}$ takes as input both the features set $X$ and the adjusted commercial price. 

The primary objective is to maximize the Global Written Margin (GWM), defined as the total sum of the margin weighted by conversion. This serves as a significant indicator of profitability.  In addition, two local constraints allows to maintain that individual coefficient in a specific range  $c_i \in [a,b]$. These constraints ensure that pricing strategies adhere to a realistic and controllable strategy, mitigating potential risks associated with uncontrolled pricing and are also designed to avoid penalization on vulnerable segments (i.e., low price elasticity segment). The last constraint is crucial to maintain a sufficient level of conversion rate forcing the average portfolio conversion above a specific threshold $\gamma$. 

For practical reasons, a relaxation is employed. The following equation corresponds to the  linearization of the formula~\ref{eq:indopt}.

\begin{equation}
\argmax_{c_1,..,c_n} \sum_{i=1}^{n}[(c_i * h_{w_h}(x_i) - h_{w_h}(x_i)) * f_{w_f}(x_i, c_i * h_{w_h}(x_i))] + \lambda * \frac{1}{n} \sum_{i=1}^{n} f(x_i, c_i * h_{w_f}(x_i))
\label{eq:relaxind}
\end{equation}

In this formulation, the specific threshold $\gamma$ is no longer explicit. However, the hyperparameter $\lambda \in \mathbb{R^+}$ allows for a trade-off without the need for precise information about the minimum average conversion rate. Higher values of $\lambda$ lead to increased conversion rates but generally result in a lower Gross Written Margin (GWM). 

This problem is typically addressed using the Sequential Quadratic Programming method (SQP). It is important to note that local constraints can be preserved in practice by addressing discrete optimization issues, employing specific choices for pricing rate adjustments (e.g., +10\%, +30\% with 1-point fixed increments). This methodology enhances the model's flexibility and practical utility in real-world applications, albeit potentially at the cost of optimum accuracy.

\subsection{Discrimination and Unfairness}

The quest for fairness in AI systems requires the articulation of a clearly defined \emph{fairness criteria}. First, there is the concept of information sanitization that restricts the use of sensitive data in predictive model training. A key instance from the European insurance sector is the implementation of the Gender Directive. Although nearly a decade passed before legal guidelines for its application to insurance activities were established, the directive was fundamentally aimed at upholding the principle of equal treatment between men and women in the access and provision of goods and services, including insurance. The result was a prohibition on the use of gender as a rating variable in insurance pricing. The directive enforced gender equality in insurance pricing across the European Union from December 21, 2012, as detailed in~\citep{Schmeiser2014Unisex}. 

However, a major challenge for actuaries is that sensitive variables can be highly correlated with several other features. For example, the combination the driver’s occupation and car's size, color could inadvertently introduce gender bias in the prediction of car insurance prices. Geographic information has also been noted as dependent to the race or national origin in Insurance~\citep{saxena2024spatial}. An illustrative instance is shown ~\citep{angwin2017minority}, which revealed that individuals residing in neighborhoods predominantly populated by racial minorities are subject to elevated insurance premiums compared to individuals with equivalent risk profiles residing in other neighborhoods.
A significant aspect of fairness in the Fair Machine Learning community is statistical or group fairness. This partitions the world into groups defined by one or several sensitive attributes. It requires that a specific relevant statistic about the classifier is equal across those groups. This is particularly challenging in the context of continuous sensitive attributes, where it is crucial to guarantee distributional independence instead of looking at average expectation between groups. 
In the following, we outline the most popular definition and mitigating strategy used in recent research. 

\subsubsection{Demographic Parity}

The most common objective in fair machine learning is \emph{Demographic parity} by~\citep{Dwork2011}. Based on this definition, a model is considered fair if the output prediction $\widehat{Y}$ from features $X$ is independent of the sensitive attribute $S$: 
$\widehat{Y}\indep S$.





\begin{definition}
A machine learning algorithm achieves \emph{Demographic Parity} if the associated prediction $\widehat{Y}$
is independent of the sensitive attribute $S$~\footnotemark[1]: 
\begin{eqnarray}
\mathbb{P}(\widehat{Y} \leq y |S=s)=\mathbb{P}(\widehat{Y} \leq y ),~\forall s.
\label{def:demographic_parity}
\end{eqnarray}

\end{definition}
\footnotetext[1]{For the binary case, it is equivalent to $\mathbb{P}(\widehat{Y} = y |S)=\mathbb{P}(\widehat{Y} = y )$}

The use of \emph{Demographic Parity} was originally introduced in this context of binary scenarios ~\citep{Dwork2011}, 
where the underlying idea is that each demographic group has the same chance for a positive outcome.

\begin{definition}
A classifier is considered fair according to the demographic parity principle if $$\mathbb{P}(\widehat{Y}=1|S=0)=\mathbb{P}(\widehat{Y}=1|S=1).$$
\end{definition}

Considering the evaluation of fairness between the sensitive attribute $S$ and a continuous variable $P=C \times h(X)$, which represents the final commercial premium offered to customers, we delve into the continuous scenario of Demographic Parity. Traditional methods for measuring dependence in continuous cases include Pearson's correlation, Kendall's tau, and Spearman's rank correlation. However, these measures predominantly capture specific types of association patterns, such as linear or monotonically increasing relationships. For example, two variables that have a quadratic relationship, 
would not exhibit correlation in the Pearson sense. 
To overcome the limitations of these conventional linearity-centric measures, the HGR (Hirschfeld-Gebelein-Rényi) coefficient offers a viable alternative. It is a normalized measure which is capable of correctly measuring linear and non-linear relationships, it can handle multi-dimensional random variables and it is invariant with respect to changes in marginal distributions~\citep{lopez2013randomized}.
\footnotetext[1]{$\rho(U, V)$ := $\frac{Cov(U;V)}{\sigma_{U}\sigma_{V}}$, where $Cov(U;V)$, $\sigma_{U}$ and $\sigma_{V}$ are the covariance between $U$ and $V$, the standard deviation of $U$ and the standard deviation of $V$, respectively.}



 
\begin{definition}
For two jointly distributed random variables $U \in \mathcal{U}$ and $V \in \mathcal{V}$
, the Hirschfeld-Gebelein-R\'enyi maximal correlation is
defined as:

\begin{eqnarray}
HGR(U, V) = \sup_{\substack{ \phi:\mathcal{U}\rightarrow \mathbb{R},\psi:\mathcal{V}\rightarrow \mathbb{R}\\
           E(\phi(U))=E(\psi(V))=0 \\
           E(\phi^2(U))=E(\psi^2(V))=1}} \rho(\phi(U), \psi(V)) 
          = \sup_{\substack{ \phi:\mathcal{U}\rightarrow \mathbb{R},\psi:\mathcal{V}\rightarrow \mathbb{R}\\
           E(\phi(U))=E(\psi(V))=0 \\
           E(\phi^2(U))=E(\psi^2(V))=1}} E(\phi(U)\psi(V))
\label{hgr}
\end{eqnarray}

where $\rho$ is the Pearson linear correlation coefficient~\footnotemark[1] with some measurable functions $\phi$ and $\psi$. 
\end{definition}

The HGR coefficient is equal to 0 if the two random variables are independent. If they are strictly dependent the value is 1.  
The dimensional spaces for the functions $\phi$ and $\psi$ are infinite. This property is the reason why the HGR coefficient proved difficult to compute. One way  to approximate this coefficient is to require that $\phi$ and $\psi$ belong to Reproducing Hilbert Kernel's spaces (RKHS) by taking the largest canonical correlation between two sets of copula random projections. This has been done efficiently under the name of Randomized Dependency Coefficient (RDC)~\citep{lopez2013randomized}. We will make use of this approximated metric. 


\subsubsection{Using Adversarial Learning to Ensure Fairness}\label{sec:notations:method}

Machine learning fairness interventions can be broadly classified into three main categories: pre-processing, in-processing, and post-processing. Pre-processing techniques~\citep{kamiran2012data,bellamy2018ai,calmon2017optimized} seek to adjust the input data to reduce bias prior to training. In contrast, post-processing methods~\citep{hardt2016equality,chen2019fairness} aim to correct the outputs of already trained models. Meanwhile, in-processing strategies~\citep{Zafar2017mechanisms,zhang2018mitigating,wadsworth2018achieving,louppe2017learning} directly tackle bias during the model training phase. This paper places an emphasis on in-processing fairness, with a particular focus on adversarial learning  as this approach is recognized as an effective framework for scenarios where intervention during the training process is viable~\citep{louppe2017learning,wadsworth2018achieving,zhang2018,grari2019fairness}.

Among the notable fair adversarial methods, the approach developed by \citet{zhang2018} is highlighted:

\begin{equation}
    \min_{w_g} \quad 
    {\mathbb{E}_{(x,y,s) \sim p}{\;\mathcal{L_Y}(g_{w_g}(x), y)}} \\
    \quad 
    \textrm{s.t.} \quad 
    \min_{w_a} \mathbb{E}_{(x,y,s) \sim p}{\mathcal{L_S}(a_{w_a}(h_{w_h}(x)), s)} > \epsilon'
    \label{eq:global-fairness-adv}
\end{equation}

In this adversarial setup, $\mathcal{L}_\mathcal{Y}$ denotes the loss associated to the predictor objective, $\mathcal{L}_\mathcal{S}$ denotes the loss associated with the sensitive attribute reconstruction, such as a log loss for a binary sensitive attribute. The objective is to train a model, $g_{w_w}$, that not only minimizes the conventional loss associated with the prediction task but also ensures that an adversarial model, $a_{w_a}$ with parameters $w_a$, is unable to accurately infer the sensitive demographic groups from the predictor's output, $g_{w_g}(x)$. Specifically, the loss $\mathcal{L_S}(a_{w_a}(g_{w_g}(x)), s)$ should exceed a predefined threshold $\epsilon'$. To effectively balance the predictor's accuracy and the adversary's inability to determine sensitive attributes, a relaxed version of the formulation is adopted: $\min_{w_g}\max_{w_a} \mathbb{E}_{(x,y,s) \sim p}{\left[\mathcal{L_Y}(g_{w_g}(x), y)\right]} - \lambda_S \mathbb{E}_{(x,y,s) \sim p}{\left[\mathcal{L_S}(a_{w_a}(g_{w_g}(x)), s)\right]}$. Here, the coefficient $\lambda_S \in \mathbb{R}^+$ modulates the balance between the predictor's accuracy and the adversary's performance on reconstructing the sensitive attribute. A higher $\lambda_S$ value intensifies the focus on restricting the adversary’s capability to predict $S$, while a lower value favors the enhancement of the predictor’s efficiency in its main predictive task.


However, our consideration involves debiasing a commercial predictor model characterized by continuous outputs. The application of these debiasing algorithms within regression tasks, such as those using mean square loss, fails to achieve the demographic parity objective - it only aligns the conditional expectation $\mathbb{E}(S \vert \widehat{Y})=\mathbb{E}(S)$ rather than ensuring probability independence. 
For this continuous problem, the $HGR$ approach proposed by~\citep{grari2019fairness} uses an adversarial network that takes the form of two inter-connected neural networks for approximating the optimal transformations functions $\phi$ and $\psi$. The HGR estimation denoted as $\widehat{HGR}^{w_\phi,w_\psi}_{U\sim\mathcal{D_U},V\sim\mathcal{D_V}}(U, V)$ is the HGR neural estimation  between two variables $U$ and $V$, computed via two inter-connected neural networks $\phi$ and $\psi$ with parameters $w_\phi$ and $w_\psi$ ~\citep{grari2019fairness,grari2021learning}:
\begin{equation}
\label{eq:HGR_NNa}
\mathop{\widehat{HGR}^{w_\phi,w_\psi}}_{U\sim\mathcal{D_U},V\sim\mathcal{D_V}}(U, V) =\mathop{max}_{w_\phi,w_\psi}\mathbb{E}_{U\sim\mathcal{D_U},V\sim\mathcal{D_V}}(\widehat{\phi}_{w_{\phi}}(U)\widehat{\psi}_{w_{\psi}}(V))
\end{equation}
where $\mathcal{D_U}$ (resp. $\mathcal{D_V}$) is the distribution of $U$ (resp. $V$), and $\widehat{\phi}$ (resp. $\widehat{\psi}$) refer to standardized outputs of network $\phi$ (resp. $\psi$). The mitigation approach proposed uses an adversarial network to penalize this HGR estimation:
$\argmin_{w_g}\max_{{w_{\phi},w_{\psi}}}\mathbb{E}_{(x,y,s) \sim p}\mathcal{\mathcal{L_Y}}(g_{w_g}(x),y)
    + \mathop{\widehat{HGR}^{w_\phi,w_\psi}}_{(x,s)\sim p}(g_{w_g}(x), s) 
$


In this paper so far, we have primarily focused on mitigating biases of predictor model with conventional loss functions $\mathcal{L}_Y$, such as the log loss function or mean squared error (MSE). 
It's worth noting the absence of a loss function directly tailored to commercial premiums. For example, Equation \eqref{eq:indopt} has not been treated as a differentiable loss function in current state-of-the-art. Traditionally, due to the intricate nature of commercial premiums, most approaches concentrate on debiasing the average cost and or frequency— pure premium~\citep{lindholm2022discrimination,grari2022fair,xin2023antidiscrimination,lindholm2022discussion,lindholm2023fair,hu2023fairness,moriah2023measuring}—as it effectively mitigates biases inherent in traditional loss functions and can be readily applied at the pure premium layer.

However, complexities arise when considering commercial premium pricing. The introduction of a secondary layer of  commercial premium may potentially compromise the fairness achieved at the pure premium level. This introduces an additional objective balancing profit margins with conversion rates, which, inadvertently, could reintroduce biases.

\subsection{Fairness Challenges in Commercial Premium}

This section delves into the complexities associated with integrating fairness considerations into commercial premium optimization, with a specific focus on car insurance pricing. To empirically investigate these challenges, we conducted experiments utilizing the Atoti Dataset\footnote{Further details provided in the Experimental Section}, specifically designed for elasticity optimization purposes. In this section, we 
perform an Indirect Ratebook Optimization with an XGBoost model trained on individually optimized prices, wherein we adjusted the hyperparameter $\lambda$ in Eq.~\ref{eq:relaxind}, 
to find a balance between Gross Written Margin (GWM) and minimum conversion rates.

\begin{figure}[ht!]
    \centering
    \includegraphics[width=0.49\linewidth]{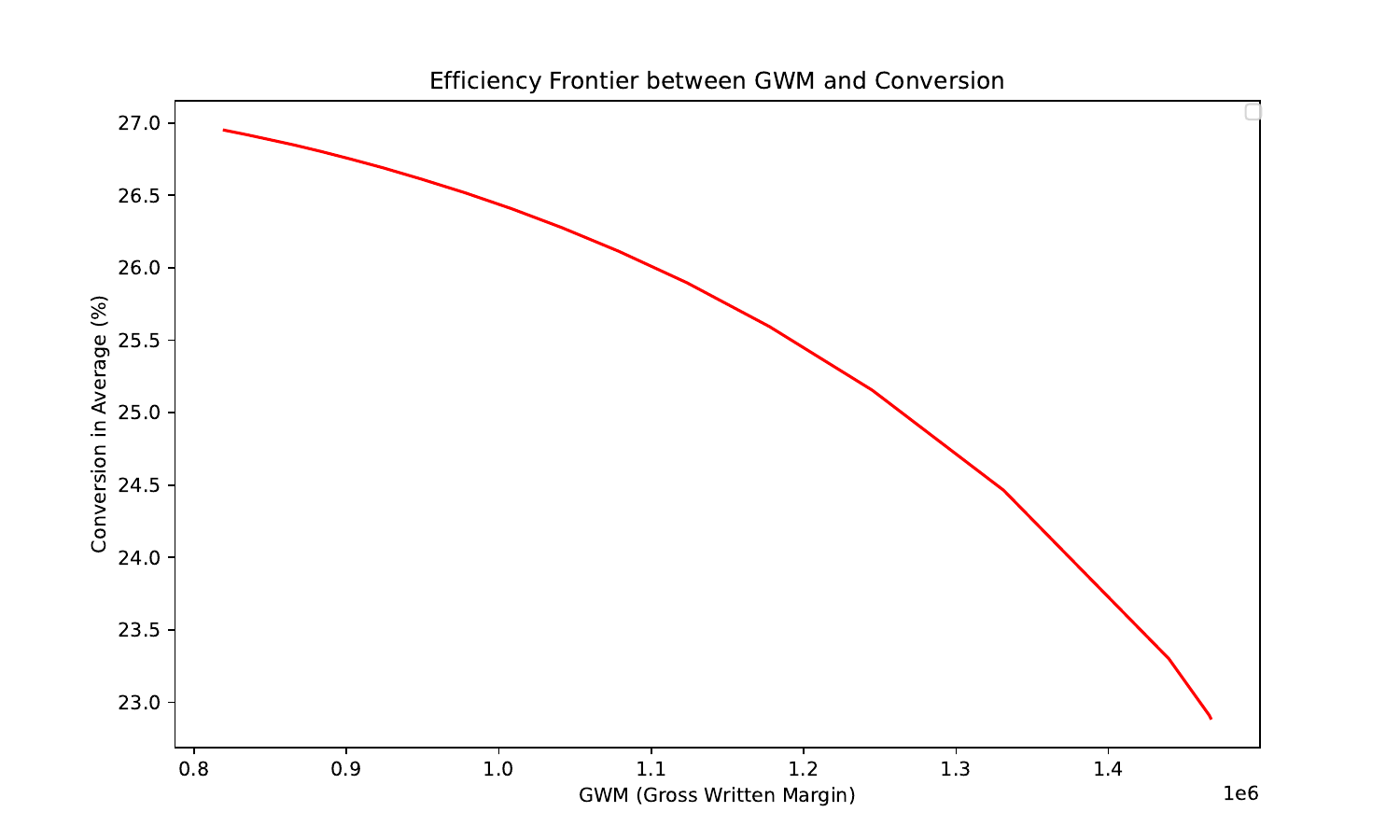}
    \includegraphics[width=0.49\linewidth]{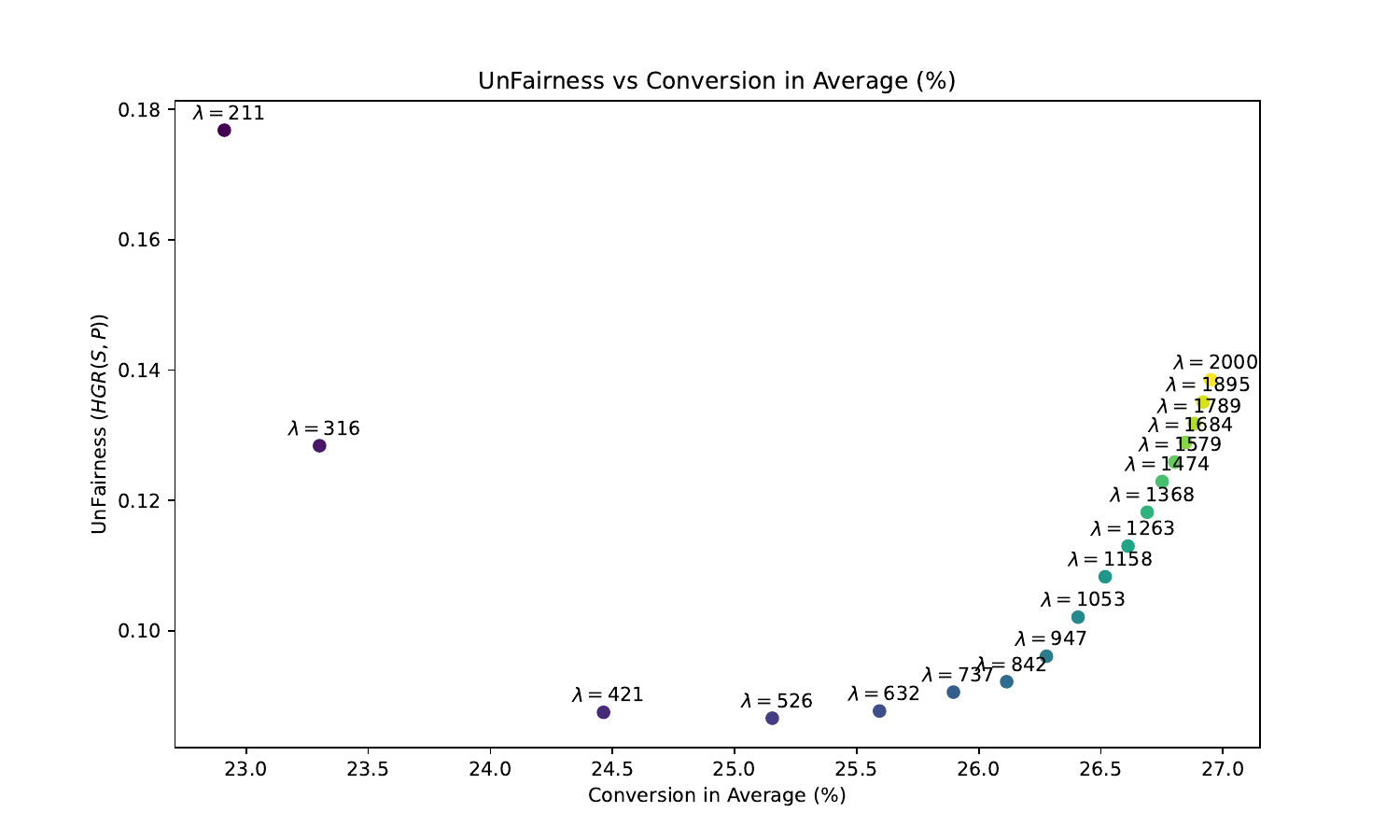}
    \caption{Efficiency frontier analysis. 
    }
    \label{fig:results-fairnesscomplexity}
\end{figure}

We explore various scenarios by incrementally adjusting the $\lambda$ hyperparameter and evaluating the resultant fairness levels, measured with respect to a sensitive attribute—\emph{age}, which was excluded from the model training but utilized solely for fairness assessment. Our findings, depicted in Figure~\ref{fig:results-fairnesscomplexity}, indicate that $\lambda$ the influences fairness. For example in extreme settings, characterized by either low conversion rates (approximately 23\%) or higher conversion rate (approximately 27\%)
—exhibit the most pronounced biases. Interestingly, bias decreases at intermediate $\lambda$ values, highlighting the intricate interplay between fairness considerations and premium pricing optimization. These results suggest that addressing fairness in commercial premium  introduces a layer of complexity extending beyond the simple management of profit or conversion rates.

\section{\emph{OptiGrad}: Price Elasticity Optimization via Gradient Descent} 

Price elasticity optimization poses a challenge for traditional methods to enforce fairness due to the presence of non-differentiable elements. To overcome this obstacle, we propose a novel approach called \emph{OptiGrad}. This method introduces a differentiable coefficient model, denoted as $c_{w_c}$ (with parameters $w_c$) that serves as the calculation of the commercial premium $p_i=c_{w_c}(x_i) * h_{w_h}(x_i)$, which takes customer features $X$ as input. 

In the following, we will explore two distinct implementations of \emph{OptiGrad}. The first implementation, discussed in detail in Subsection~\ref{sec:methodprinc}, initially focuses on the optimization process without considering fairness enforcement. Subsection~\ref{sec:fairmethod} incorporates  fairness into the methodology. 

\subsection{Formalization without Fairness}\label{sec:methodprinc}

In this section, we consider the actuarial pure premium model $h_{w_h}$ and the conversion model $f_{w_f}$ to be \emph{differentiable} and already trained. Both models are typically represented using General Linear Models (GLMs), which is a common practice in the actuarial field, although it is worth noting that they can also be represented as neural networks.

\paragraph{\emph{OptiGrad} Formulation}

Our main proposition introduces a differentiable predictor model ${c}_{w_c}$ designed to maximize profit margins while ensuring a minimum conversion rate above a level $\gamma$. This formulation is a Direct Ratebook optimization  which determines the optimal weight $w_c$ of the coefficient ${c}_{w_c}$. 
The mathematical formulation is the following:

\begin{equation}
\begin{aligned}
\max_{w_c} \quad & {\mathbb{E}_{x \sim p}[{\;(c_{w_c}(x) * h_{w_h}(x)-h_{w_h}(x))*f_{w_f}(x,c_{w_c}(x) * h_{w_h}(x)) }}] \\
\textrm{s.t. } &c_{w_c}(x) \geq a \\
              &c_{w_c}(x) \leq b \\
              &\mathbb{E}_{x \sim p}[{f_{w_f}(x,c_{w_c}(x) * h_{w_h}(x)) }]>\gamma     
\end{aligned}
\label{eq:Elastic_GradFormulation}\end{equation}

In the above equation, $a$ and $b$ denote the lower and upper bounds for coefficient $c_{w_c}(x)$, respectively, thus forming the interval $[a, b]$. $\gamma$ represents the minimum expected conversion rate $f_{w_f}(x,c_{w_c}(x) * h_{w_h}(x))$ over the distribution $p$. 

\paragraph{Bounded Interval Constraint} The primary challenge in this formulation arises from the necessity to confine the coefficient 
within the bounded interval $[a, b]$. To address this constraint while circumventing penalization inefficiencies, we employ a transformed sigmoid monotonic function, defined as $\widehat{c}_{w_c}(x)=\sigma(c_{w_c}(x)) \cdot (b-a)+a$. This approach ensures the monotonicity and differentiability 
while bounding its output within the interval $[a,b]$, where $\widehat{c}_{w_c}:\mathcal{X} \rightarrow [a,b]$.



To make Eq.~\ref{eq:Elastic_GradFormulation} differentiable, we implement a Lagrangian relaxation, introducing the $\lambda_f$ hyperparameter. The overall \emph{OptiGrad} optimization problem can thus finally be formulated as:

\begin{eqnarray}
\min_{w_c} 
- \frac{1}{n} \sum_{i=1}^{n} [(\widehat{c}_{w_c}(x_i) h_{w_h}(x_i)-h_{w_h}(x_i))*f_{w_f}(x_i,\widehat{c}_{w_c}(x_i)\cdot h_{w_h}(x_i))] \label{eq:OptiGrad2} \\ - \underbrace{\lambda_f \frac{1}{n}  \sum_{i=1}^{n} {f_{w_f}(x_i,\widehat{c}_{w_c}(x_i) \cdot h_{w_h}(x_i)) }}_{\text{Maintaining a conversion rate}} \nonumber
\end{eqnarray}

The optimization problem is therefore defined as minimizing the inverse sign of Equation~\ref{eq:Elastic_GradFormulation}, which consists of maximizing the two main components: profitability (GWM) and minimum conversion under the $\lambda_f$ term, emphasizes the importance of maintaining conversion rate. 

To address this optimization challenge, we introduce the \emph{OptiGrad} algorithm (Algorithm~\ref{alg:OptiGrad}). This algorithm employs gradient descent to iteratively updates the weights \(w_c\), aiming to minimize the objective function detailed in Equation~\ref{eq:OptiGrad2}. In each iteration, the coefficient \(\widehat{c}_{w_c}\) are updated on a batch of size $b$, based on the gradient of the objective function with respect to \(w_c\). This procedure is repeated for a predetermined number of epochs or until convergence is achieved, resulting in optimized coefficients \(w_c\) that effectively balance profit maximization with adherence to minimum conversion rate requirements.


\begin{algorithm}
\caption{\emph{OptiGrad}: Price Optimization Elasticity with Gradient Descent}
\label{alg:OptiGrad}
\begin{algorithmic}[1]
\REQUIRE Training data $X$, initialize coefficients $w_c$, already trained $w_h$, $w_f$ parameters, batch size $b$, number of epochs $n_e$, learning rates $\alpha_c$, regularization parameter $\lambda_f$
\FOR{epoch $ \in [1,... n_e]$}
\FOR{each mini-batch $(X_{\text{batch}})$ of size $b$}
\STATE Update $\widehat{c}_{w_c}(X_{\text{batch}})$ using gradient descent:
\STATE \hspace{\algorithmicindent} $w_c \gets w_c - \alpha_c \nabla_{w_c} ( -\frac{1}{b} \sum_{i=1}^{b} [(\widehat{c}_{w_c}(x_i) \cdot h_{w_h}(x_i) - h_{w_h}(x_i)) \cdot f_{w_f}(x_i,\widehat{c}_{w_c}(x_i) \cdot h_{w_h}(x_i))]$
\\
\STATE  \hspace{50 mm} $- \lambda_f \frac{1}{b} \sum_{i=1}^{b} f_{w_f}(x_i,\widehat{c}_{w_c}(x_i) \cdot h_{w_h}(x_i))$
\ENDFOR
\ENDFOR
\end{algorithmic}
\end{algorithm}

\subsection{Formalization with Fairness }\label{sec:fairmethod}


In this section, the \emph{OptiGrad} formulation (referenced in Equation~\ref{eq:Elastic_GradFormulation}) is extended to integrate fairness criteria by leveraging the Hirschfeld-Gebelein-Rényi (HGR) Neural Network (HGR\_NN) architecture, referenced in section~\ref{sec:notations:method}. The choice of the HGR\_NN architecture is motivated by its capability to effectively measure dependencies with continuous features, a crucial aspect given that commercial pricing is continuous~\citep{grari2019fairness}. Integrating an HGR-based differentiable estimation into the framework enables the optimization of three key objectives: maximizing profit margins, guaranteeing a minimum conversion rates, and reducing bias.


The HGR estimation, implemented via two interconnected networks, $\phi_{w_\phi}:\mathcal{P}\rightarrow \mathbb{R}$ and $\psi_{w_\psi}:\mathcal{S}\rightarrow \mathbb{R}$, quantifies the nonlinear dependence between the adjusted commercial premium, denoted as $c_{w_c}(x) * h_{w_h}(x)$ (the input for $\phi_{w_\phi}$), and the sensitive attribute (the input for $\psi_{w_\psi}$). This framework enables the evaluation of fairness at each stage of the training process, facilitating corrective measures to mitigate bias. The goal is to ensure that the pricing strategy optimization does not inadvertently reinforce bias described by 
the sensitive attribute $S$ (binary or continuous). 

\begin{equation}
\begin{aligned}
\max_{w_c} \quad & {\mathbb{E}_{x \sim p}[{(c_{w_c}(x) *h_{w_h}(x)-h_{w_h}(x))*f_{w_f}(x,c_{w_c}(x) * h_{w_h}(x)) }]} \\
\textrm{s.t.} & c_{w_c}(x) \geq a \\
& c_{w_c}(x) \leq b \\
& \mathbb{E}_{x \sim p}[{f_{w_f}(x,c_{w_c}(x) * h_{w_h}(x)) }] > \gamma \\
& \underbrace{{\mathbb{E}_{(x,s) \sim p}(\widehat{\phi}_{w_{\phi*}}(c_{w_c}(x) * h_{w_h}(x))\widehat{\psi}_{w_{\psi*}}(s))}}_{\text{HGR Component}} < \epsilon'
\end{aligned}
\label{eq:Elastic_GradFormulationFair}
\end{equation}

Equation~\ref{eq:Elastic_GradFormulationFair} introduces the HGR component, 
it aims to ensure that the non linear dependence  between the commercial price and the sensitive attribute $S$ is below a certain threshold  $\epsilon' \in \mathbb{R}$. This HGR estimation corresponds to the expectation of the product of the optimally standardized outputs of both networks ($\widehat{\phi}_{w_{\phi^*}}$ and $\widehat{\psi}_{w_{\psi^*}}$) with $w_\phi^*$ and $w_\psi^*$ the optimal neural networks parameters maximizing the expectation of the HGR.  

By maintaining the same Lagrangian relaxation approach, the overall optimization problem of our Fair Price Elastic Optimization framework (Fair-\emph{OptiGrad}) can thus finally be formulated as:

\begin{eqnarray}\label{eq:OptiGradFair}
\min_{w_c} \max_{w_\phi,w_\psi} 
- \frac{1}{n} \sum_{i=1}^{n} [(\widehat{c}_{w_c}(x_i) *h_{w_h}(x_i)-h_{w_h}(x_i))*f_{w_f}(x_i,\widehat{c}_{w_c}(x_i)* h_{w_h}(x_i))]  \nonumber  \\ - \lambda_f \frac{1}{n}  \sum_{i=1}^{n} {f_{w_f}(x_i,\widehat{c}_{w_c}(x_i) * h_{w_h}(x_i)) }   \\
+ \lambda_S\frac{1}{n}  \sum_{i=1}^{n} \widehat{\phi}_{w_{\phi}}(\widehat{c}_{w_c}(x_i) * h_{w_h}(x_i))*\widehat{\psi}_{w_{\psi}}(s)
\nonumber 
\end{eqnarray}

In this objective, the third term represents the fairness component.  The $\lambda_S$ hyper-parameter controls its impact in the optimization. The optimization is transformed in a min-max objective in order to recover at each step the optimal neural network $\widehat{\psi}_{w_{\psi}}$ and $\widehat{\phi}_{w_{\phi}}$. This maximization of the HGR estimation can be updated by multiple steps of gradient ascent for each gradient descent iteration on $w_c$ for accurately estimating the HGR. This allows to evaluate the fairness more accurately at each stage of the training process.



\begin{algorithm}
\caption{Fair-\emph{OptiGrad}: Price Optimization Elasticity with Demographic Parity}
\label{alg:fair_OptiGrad}
\begin{algorithmic}[1]
\REQUIRE Training data $X$, initialize coefficients $w_c$, already trained $w_h$, $w_f$ parameters, initialize $\widehat{\phi}_{w_{\phi}}$ and $\widehat{\psi}_{w_{\psi}}$ networks, batch size $b$, number of epochs $n_e$, learning rates $\alpha_c$, $\alpha_{\phi}$, $\alpha_{\psi}$, regularization parameters $\lambda_f$, $\lambda_S$, number of ascent steps $n_a$
\FOR{epoch $ \in [1,..., n_e]$}
\FOR{each mini-batch $(X_{\text{batch}}, S_{\text{batch}})$ of size $b$}
\STATE Standardize the function $\widehat{\phi}_{w_{\phi}}$ and $\widehat{\psi}_{w_{\psi}}$ 

\FOR{$j = 1$ to $n_a$}
 
\STATE Update $\widehat{\phi}_{w_{\phi}}$ and $\widehat{\psi}_{w_{\psi}}$ by gradient ascent to maximize the HGR estimation
\STATE \hspace{\algorithmicindent} $w_{\phi} \gets w_{\phi} + \alpha_{\phi} \nabla_{w_{\phi}} \left( \frac{1}{b} \sum_{i=1}^{b} \widehat{\phi}_{w_{\phi}}(\widehat{c}_{w_c}(x_i) \cdot h_{w_h}(x_i)) \cdot \widehat{\psi}_{w_{\psi}}(s_i) \right)$
\STATE \hspace{\algorithmicindent} $w_{\psi} \gets w_{\psi} + \alpha_{\psi} \nabla_{w_{\psi}} \left( \frac{1}{b} \sum_{i=1}^{b} \widehat{\phi}_{w_{\phi}}(\widehat{c}_{w_c}(x_i) \cdot h_{w_h}(x_i)) \cdot \widehat{\psi}_{w_{\psi}}(s_i) \right)$
\ENDFOR
\STATE Update $\widehat{c}_{w_c}(X_{\text{batch}})$ using gradient descent for minimizing loss
\STATE \hspace{\algorithmicindent}
$w_c \gets w_c - \alpha_c \nabla_{w_c} 
(-\frac{1}{b} \sum_{i=1}^{b} [(\widehat{c}_{w_c}(x_i) \cdot h_{w_h}(x_i)- h_{w_h}(x_i)) \cdot f_{w_f}(x_i,\widehat{c}_{w_c}(x_i) \cdot h_{w_h}(x_i))]$
\\ \hspace{50 mm} $- \lambda_f \frac{1}{b} \sum_{i=1}^{b} f_{w_f}(x_i,\widehat{c}_{w_c}(x_i) \cdot h_{w_h}(x_i))$
\\ \hspace{50 mm}  $+ \lambda_S \frac{1}{b} \sum_{i=1}^{b} \widehat{\phi}_{w_{\phi}}(\widehat{c}_{w_c}(x_i) \cdot h_{w_h}(x_i)) \cdot \widehat{\psi}_{w_{\psi}}(s_i))$
\ENDFOR
\ENDFOR
\end{algorithmic}
\end{algorithm}

The algorithm takes as input a training set from which it samples batches of size $b$ at each iteration. At each iteration, it first standardizes the output scores of networks $\phi_{w_\phi}$ and $\psi_{w_\psi}$ to ensure 0 mean and a variance of 1 on the batch. Then it computes the adversary objective function to estimate
the $HGR$ neural estimate with $n_a$ gradient ascent iterations.
At the end of each iteration, the algorithm updates 
the prediction parameters $\omega_c$ 
by one step of gradient descent.






\section{Experiments}

\begin{figure}[ht!]
    \centering
    \includegraphics[width=0.49\linewidth]{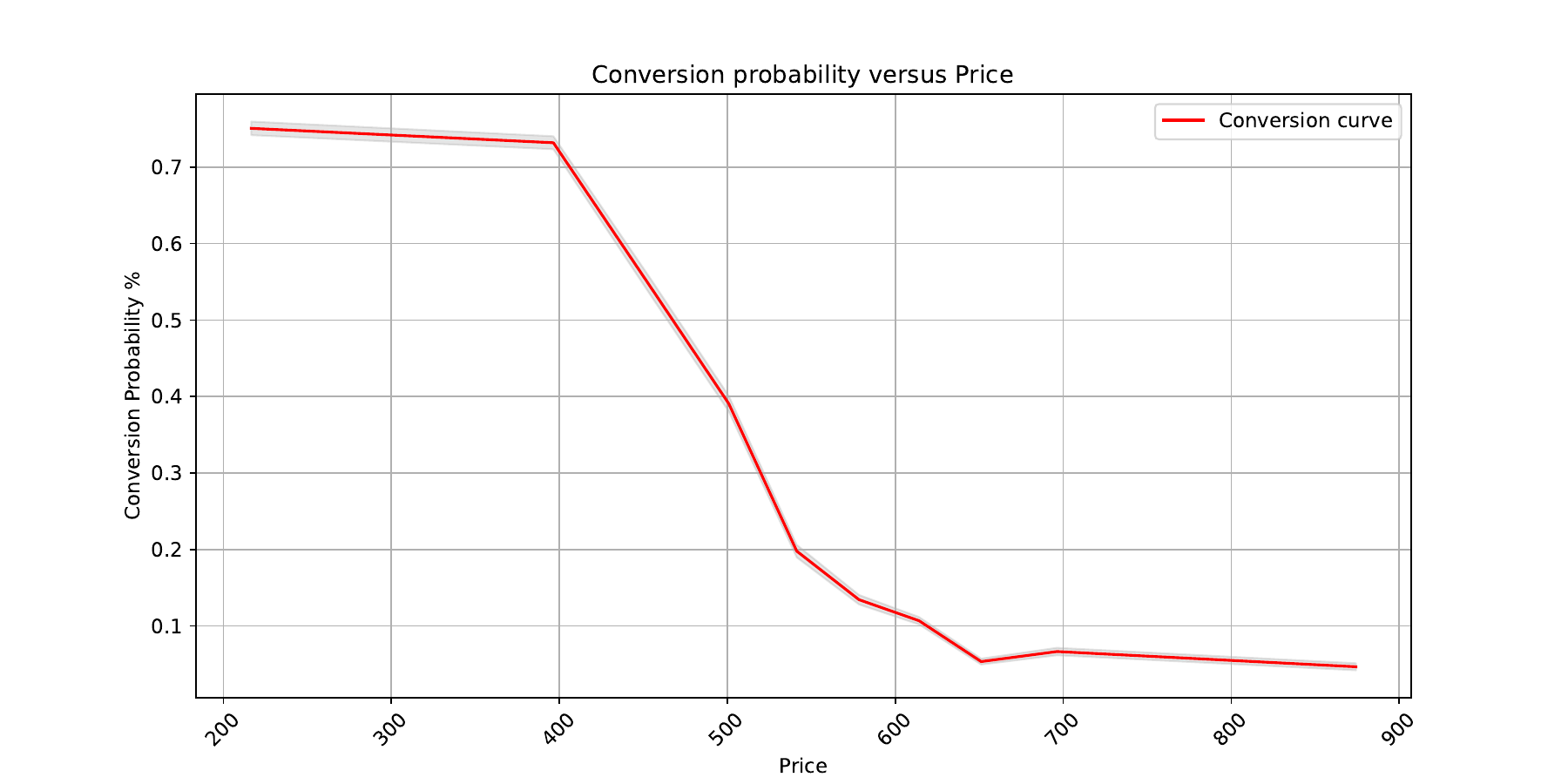}
    \caption{Conversion model exhibits a dependency on the price, with the probability decreasing as the price increases.}
    \label{fig:results-convmodel}
\end{figure}

\begin{figure}[ht!]
    \centering
    \includegraphics[width=0.49\linewidth]{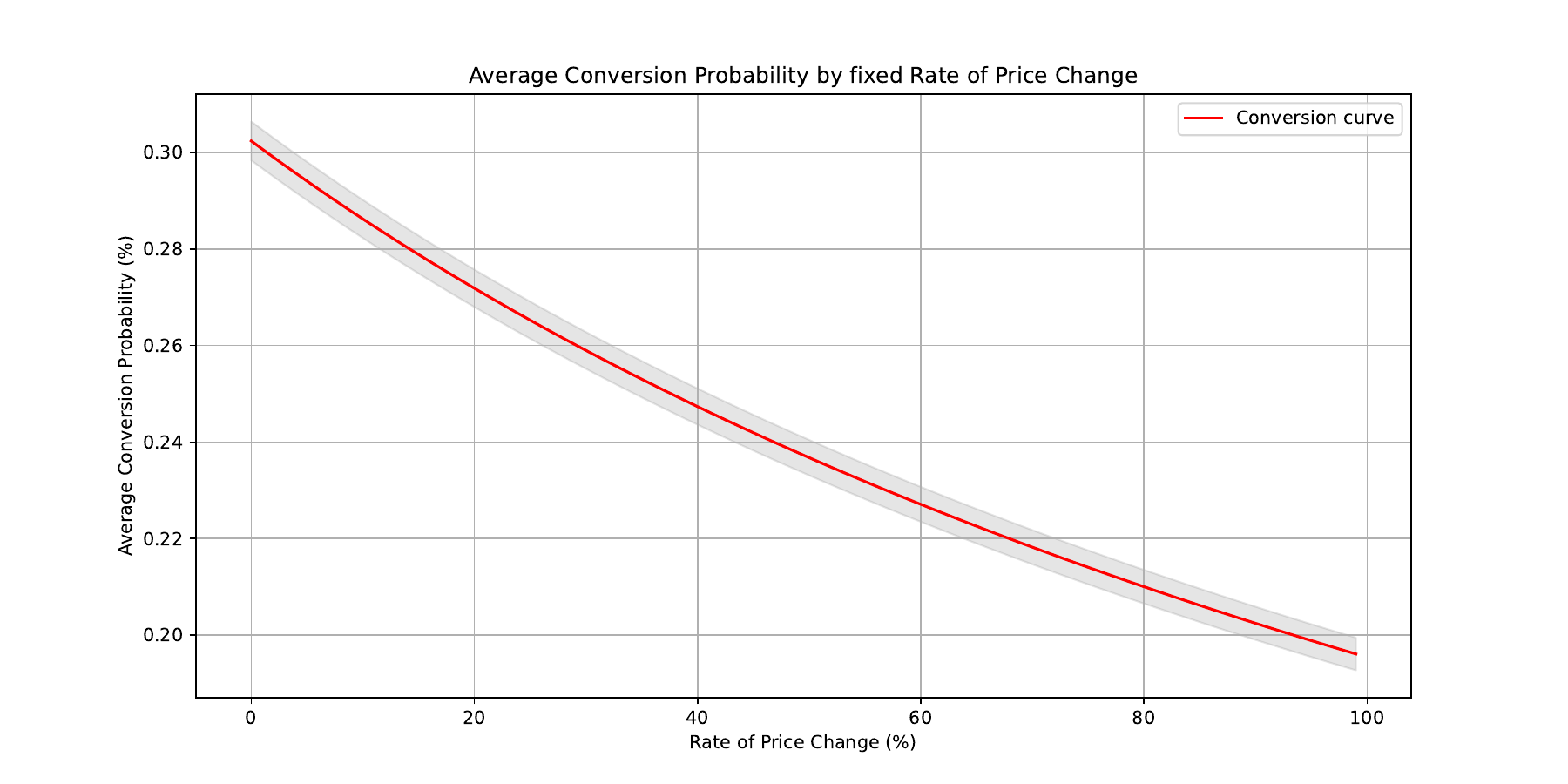}
    \caption{ Conversion rate versus a fix percentage increases of the price across the whole portfolio.}. 
    \label{fig:results-convmodel2}
\end{figure}

The dataset studied in this experiment comes from Atoti\footnote{https://data.atoti.io/notebooks/price-elasticity/data.csv \\
https://data.atoti.io/notebooks/price-elasticity/test\_df.csv} based on the Kaggle automobile insurance quotes and sales dataset\footnote{https://www.kaggle.com/datasets/ranja7/vehicle-insurance-customer-data}, enriched through synthetic data generation. It comprises 46,129 instances, divided into training (60\%), development (25\%), and test (20\%) sets. From the 18 variables present, the customer identifier ($\emph{cust\_id}$) has been excluded. The $\emph{Sale}$  variable is employed in the training of a logistic regression model, designated as $f_{w_f}(X)$. 
In line with the methodology adopted in~\citep{deLarrard}, a logarithmic transformation is applied to the historical observed price. This transformation is critical for ensuring that the conversion probability model $f_{w_f}$ outputs a value of zero for infinitely high prices and reaches its maximum at a price of zero.
As illustrated in Figure~\ref{fig:results-convmodel}, the conversion function exhibits a substantial dependency on the price, with the probability decreasing as the price increases. 
Moreover, when the price is significantly high, the demand probability is observed to approach zero. In Figure~\ref{fig:results-convmodel2}, we observe that if the price of the all portfolio are increased by a fix percentage the conversion rate decreases from $0.3$ to $0.18$.

To assess our \emph{OptiGrad} framework, two empirical scenarios have been designed. The first scenario involves the use of \emph{OptiGrad} and compares it with conventional methods that does not include fairness considerations. The second scenario focuses on enforcing demographic parity to ensure fairness.  It should be noted that although the methodology discussed in \citet{treetanthiploet2023insurance} also consider a Direct Ratebook Optimization, it cannot be directly compared to our study. The difference stems from its reliance on market quantile prices from competitors, which is outside the scope of this paper, and its omission of the local constraints that are central to our strategy.

\subsection*{\emph{OptiGrad} - Without Fairness}


In our first experiment, we compare the \emph{OptiGrad} framework against conventional state-of-the-art methods, specifically focusing on Individual Optimization (as outlined in Section~\ref{sec:indopt}), which is constrained to the training data, and an Indirect Ratebook Optimization, which employs an XGBoost model ($300$ trees with a max depth to $5$) trained on the outputs from the Individual Optimization. This offline configuration permits the evaluation of the latter on both the test and dev datasets. Moreover, for our \emph{OptiGrad} methodology we investigate the effect of model complexity—comparing a simple regression against a deep neural network—on the coefficient $c_{w_c}$. For all scenarios we have constrained the upper and lower limits to $1.6$ and $1.2$, respectively.  Each algorithm has a tradeoff hyperparameter that allows to varying the minimum conversion rate. We have therefore ensured a thorough exploration of these tradeoffs by sweeping across hyperparameter values for each algorithm.



\begin{figure}
    \centering
    \includegraphics[width=0.32\linewidth]{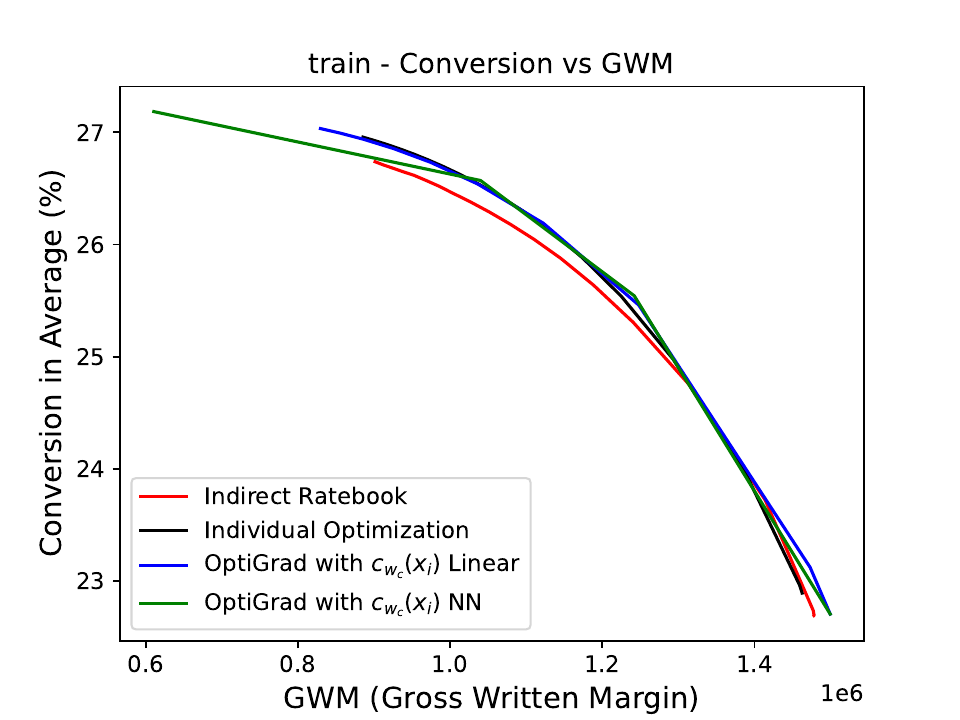}
    \includegraphics[width=0.32\linewidth]{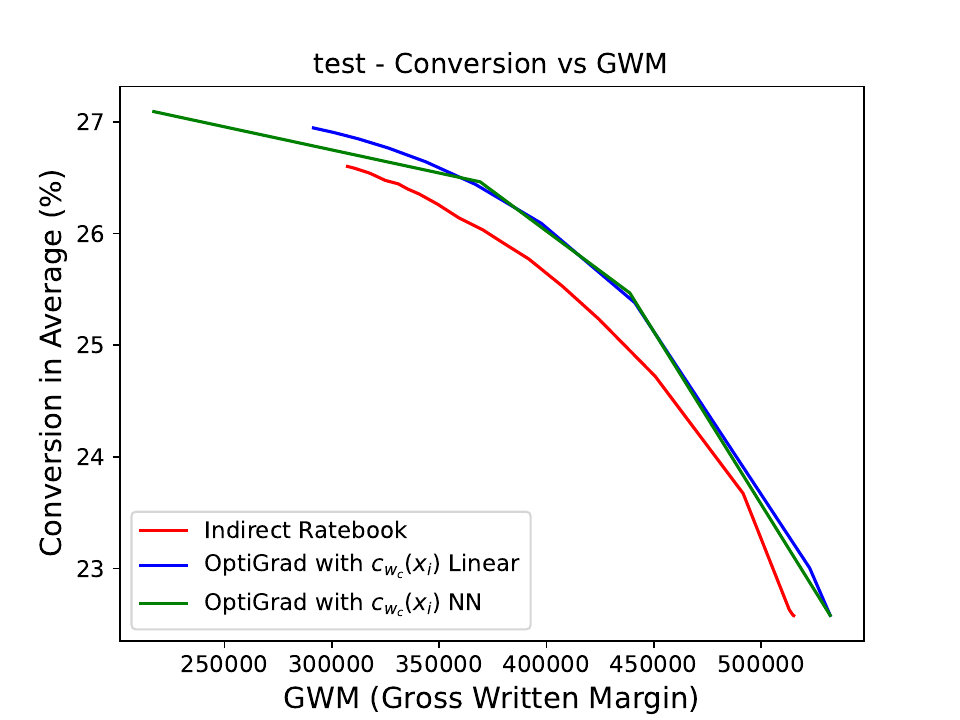}
    \includegraphics[width=0.32\linewidth]{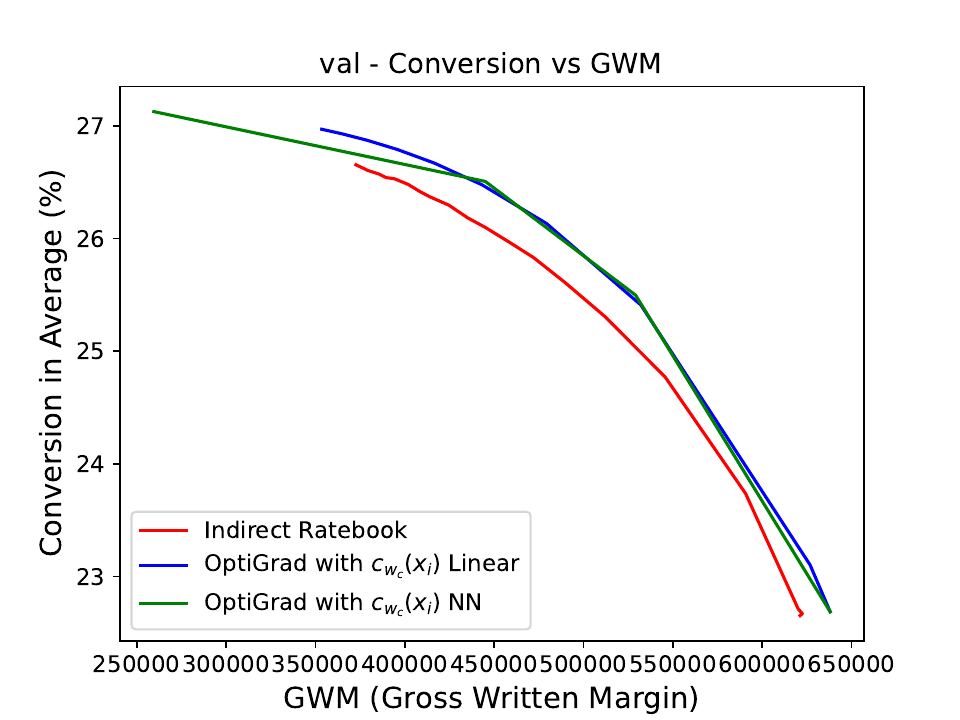}
    \caption{Efficiency frontier analysis. 
    }
    \label{fig:results-efficiencyfrontier}
\end{figure}

The results, depicted in Figure~\ref{fig:results-efficiencyfrontier}, indicate that our \emph{OptiGrad} approach outperform the traditional methodology in performance on the test and dev datasets. On the training dataset, Individual Optimization exhibits comparable effectiveness. The least effective model is the Indirect Optimization, which significantly underperformed due to the loss of performance in attempting to reverse-engineer the 
downstream individually optimized prices. Furthermore, we observe that increasing the complexity of our coefficient model $c_{w_x}$ has no impact.

\begin{figure}
    \centering
    \includegraphics[width=0.32\linewidth]{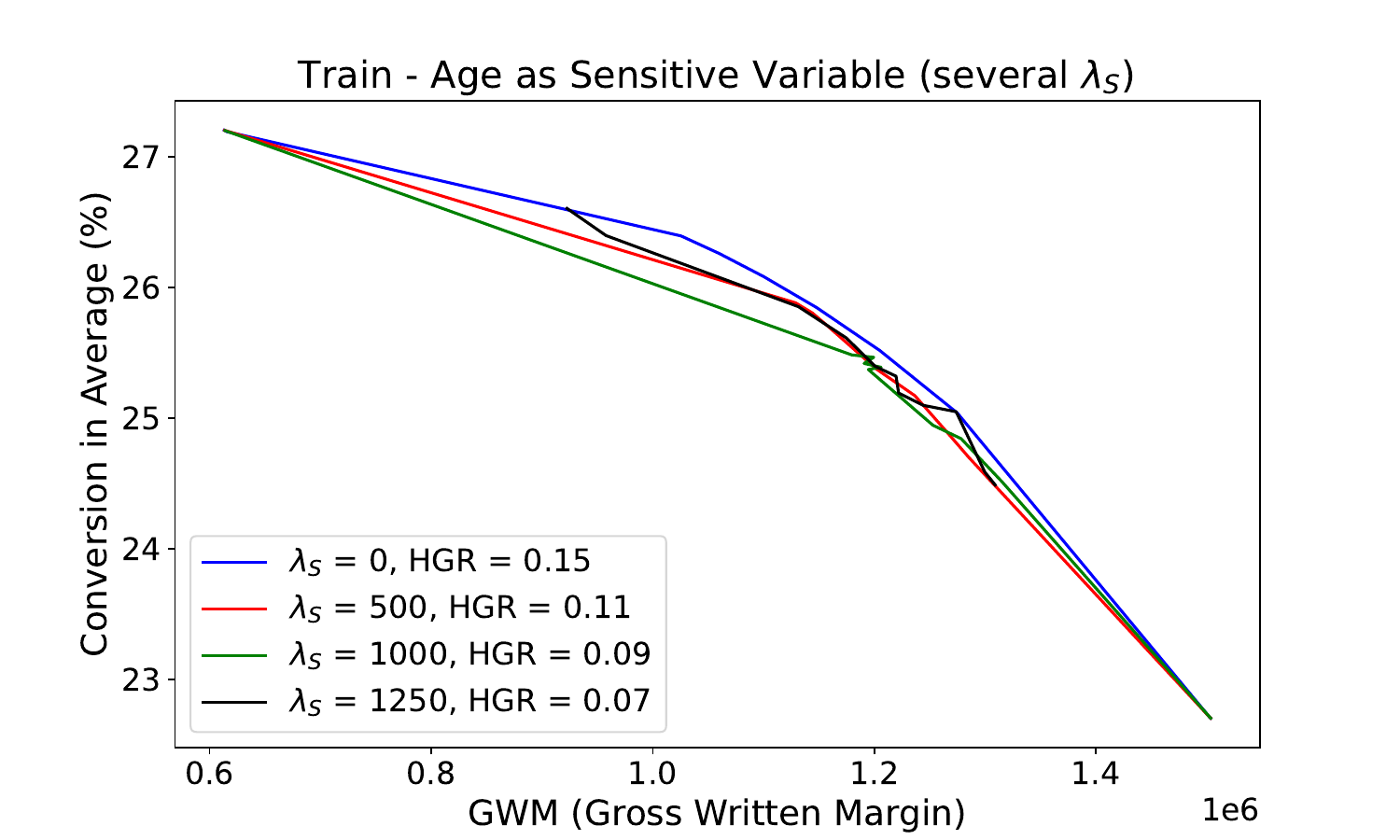}
    \includegraphics[width=0.32\linewidth]{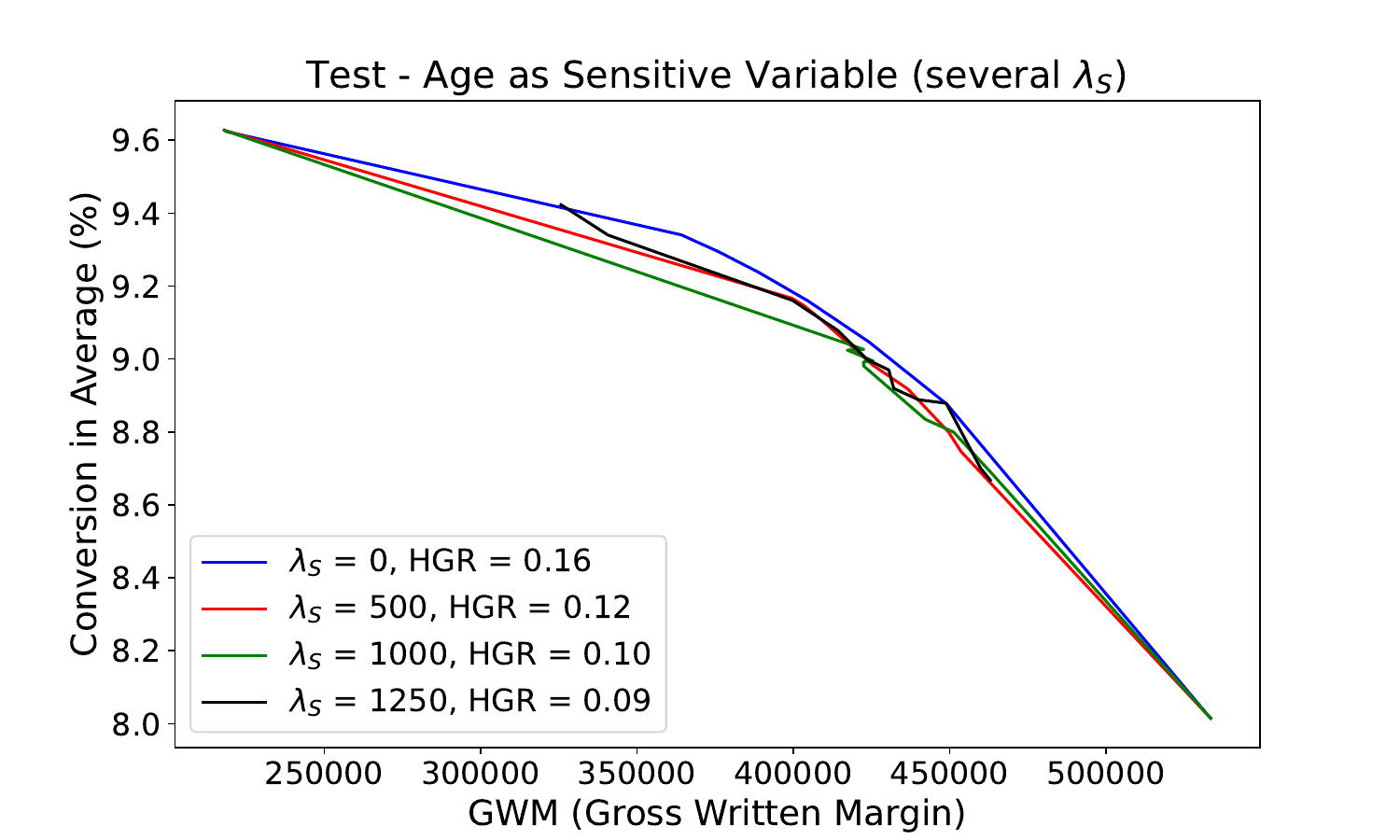}
    \includegraphics[width=0.32\linewidth]{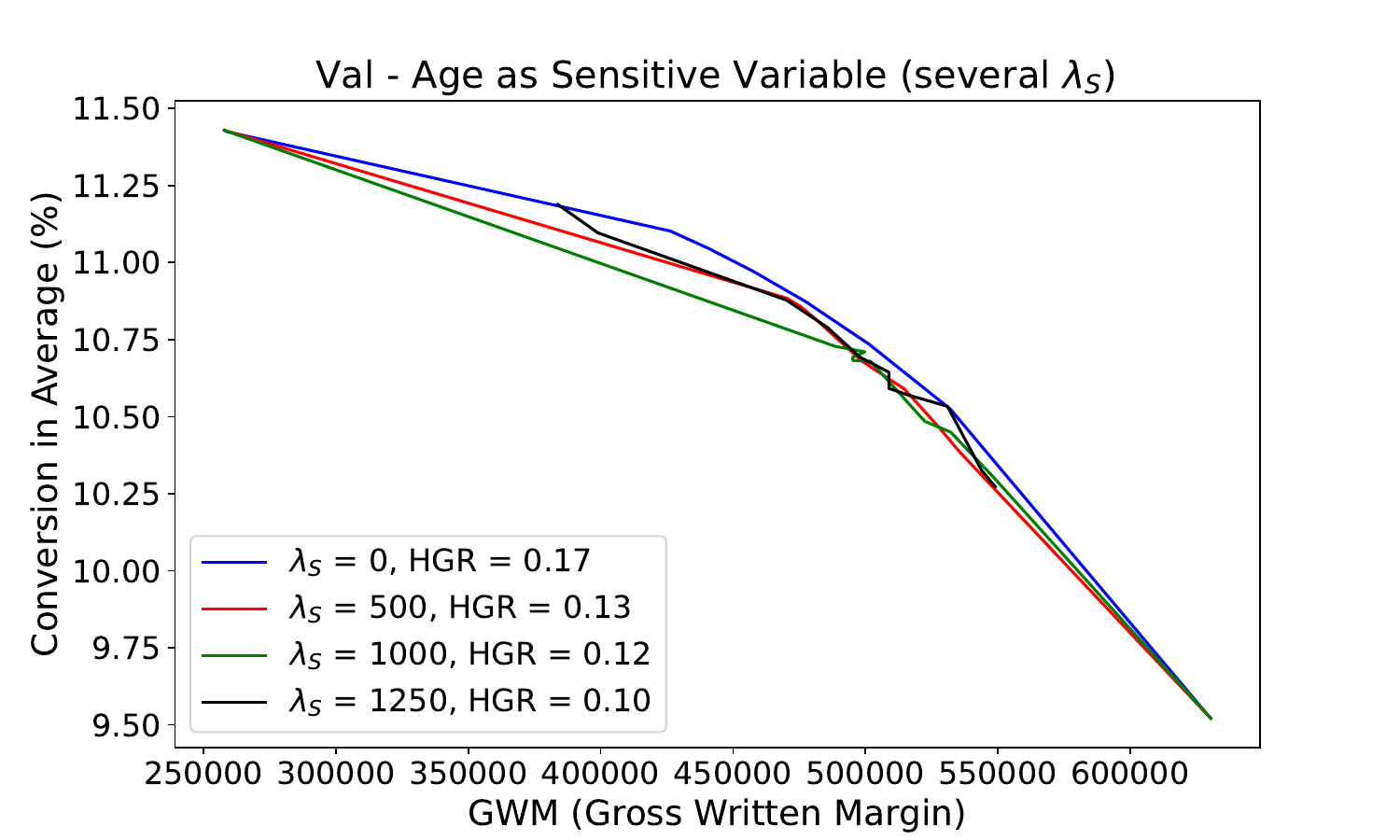} 
    \\
    \includegraphics[width=0.32\linewidth]{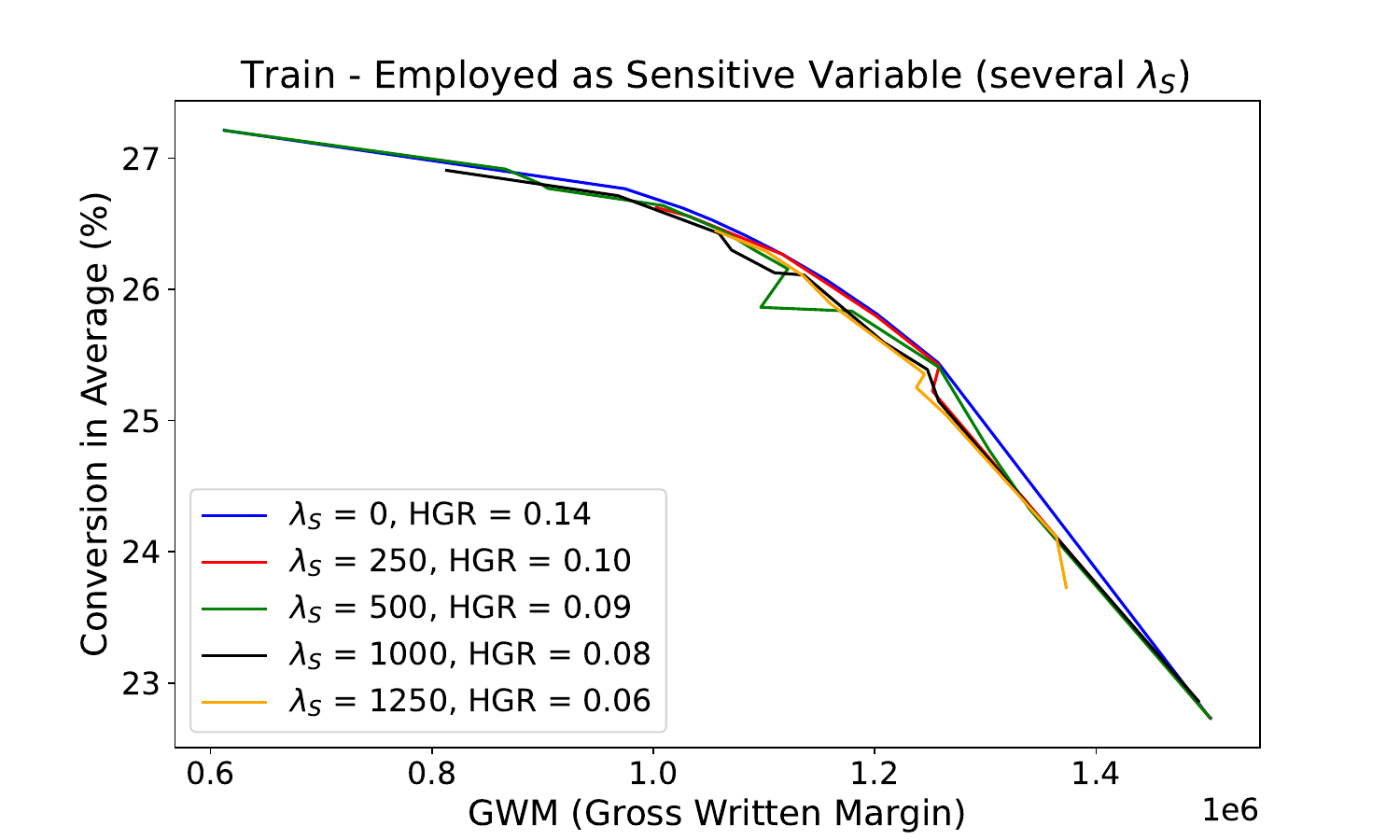}
    \includegraphics[width=0.32\linewidth]{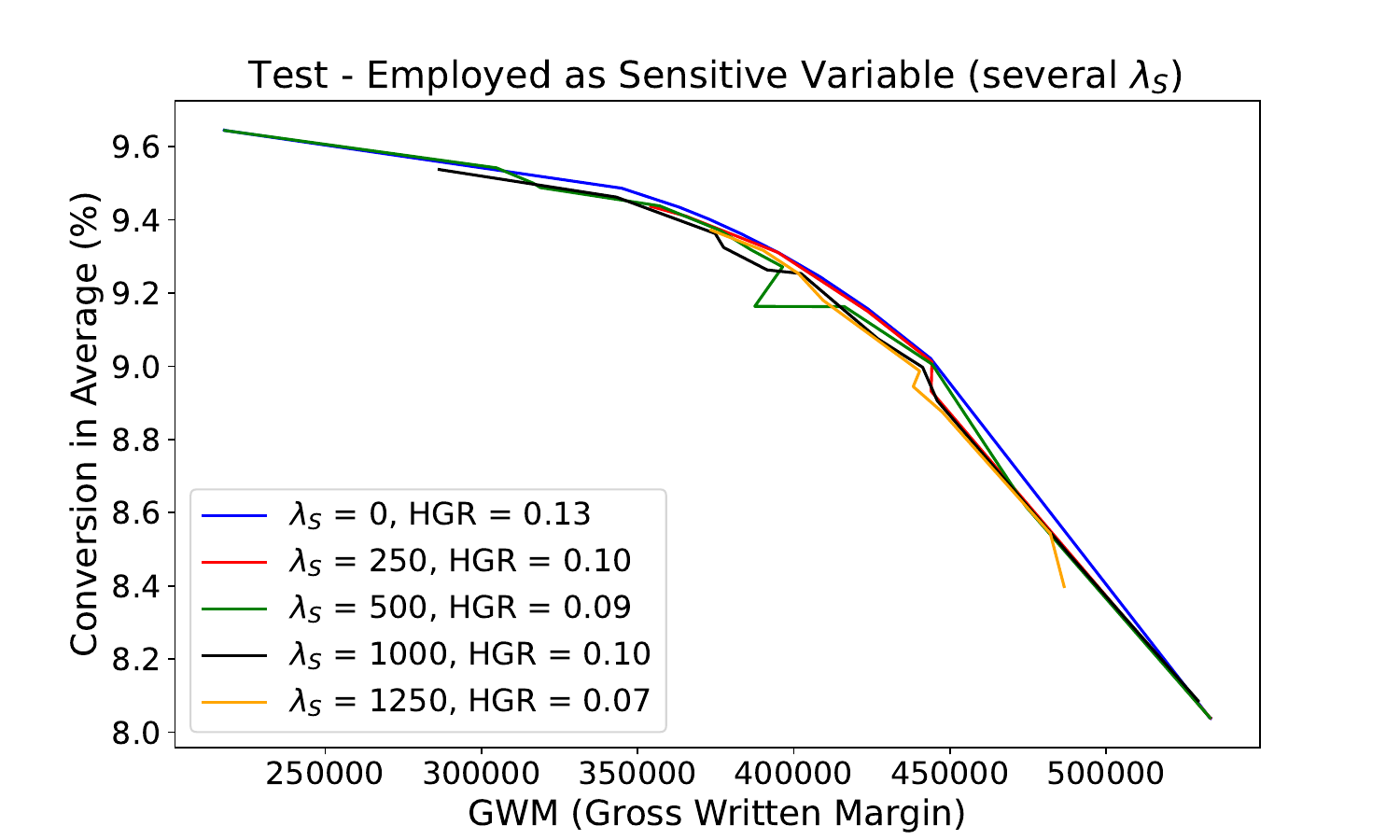}
    \includegraphics[width=0.32\linewidth]{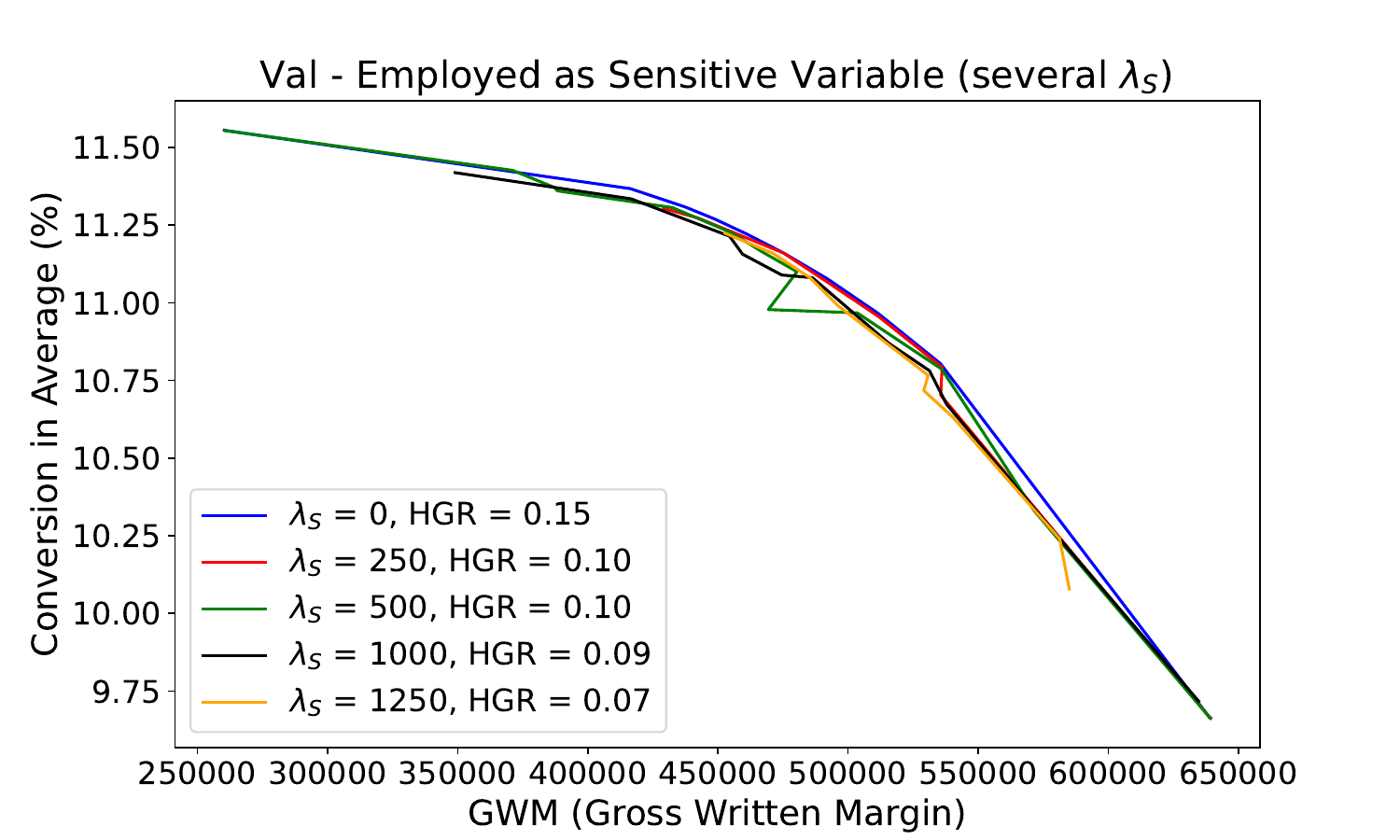} 
    \caption{Fair-\emph{OptiGrad} - Efficiency frontier analysis. 
    }
    \label{fig:results-fairness_gwmvsConv}
\end{figure}

\subsection*{\emph{OptiGrad} - With Fairness Enforcement}

To analyse the enforcement of fairness of our \emph{OptiGrad} framework, we investigated two distinct scenarios: 
(i) with a continuous sensitive attribute: Age, (ii) 
and a binary sensitive attribute: Employment status. 
It is important to note that these sensitive attributes are used solely during the training phase and are not employed in the testing phase, which is dedicated to the assessment of the 
fairness.

Our Fair-\emph{OptiGrad} algorithm incorporates two hyperparameters to balance the trade-off between achieving a minimum conversion rate and enforcing fairness. Following a similar approach to our initial scenario, we conducted an exhaustive hyperparameter sweep to optimally balance these two components.

In Figure~\ref{fig:results-fairness_gwmvsConv}, we observe that adjusting the parameter \(\lambda_S\) effectively promotes fairness across different demographics. Increasing \(\lambda_S\) from 0 to 1250 results in a significant reduction in the HGR from 0.15 to 0.07 for age, and from 0.14 to 0.06 for employment. This indicates a robust enforcement of fairness as \(\lambda_S\) increases. Most important observation is that the Pareto front for the different $\lambda_s$ remains similar with the exception of the extreme values ($\lambda_s =1250$) while unfairness (HGR) is reduced in average.

\section{Conclusion}

To conclude, our research introduces the \emph{OptiGrad} framework as an effective tool to optimize the complexities of (i) profit maximization, (ii) conversion rate optimization, and (iii) fairness in the context of commercial insurance premiums. This approach outperforms traditional discrete pricing methods by utilizing continuous rate optimization, resulting in improved precision and efficiency. Importantly, \emph{OptiGrad} innovatively integrates fairness directly into the computation of commercial premiums, addressing a blind spot in previous 
approaches that focused solely on optimizing pure premiums. 

The outcomes of this study underscore the effectiveness of traditional gradient descent techniques in augmenting insurance pricing strategies, providing a foundation for achieving a balanced between profitability and ethical considerations in insurance pricing. Nevertheless, this approach implies the differentiability of the different components.

Future research directions include better understanding extreme cases where fairness requirement may be unrealistic. Furthermore, we are also interested on the exploration of additional fairness criteria, such as equalized odds or equal opportunity. This could further refine and improve the fairness and societal impact of insurance pricing algorithms. 

\vskip 0.2in
\bibliography{sample}

\begin{thebibliography}{31}
\providecommand{\natexlab}[1]{#1}
\providecommand{\url}[1]{\texttt{#1}}
\expandafter\ifx\csname urlstyle\endcsname\relax
  \providecommand{\doi}[1]{doi: #1}\else
  \providecommand{\doi}{doi: \begingroup \urlstyle{rm}\Url}\fi

\bibitem[Angwin et~al.(2017)Angwin, Larson, Kirchner, and Mattu]{angwin2017minority}
J.~Angwin, J.~Larson, L.~Kirchner, and S.~Mattu.
\newblock Minority neighborhoods pay higher car insurance premiums than white areas with the same risk.
\newblock \emph{ProPublica, April}, 5:\penalty0 2017, 2017.

\bibitem[Bellamy et~al.(2018)Bellamy, Dey, Hind, Hoffman, Houde, Kannan, Lohia, Martino, Mehta, Mojsilovic, et~al.]{bellamy2018ai}
R.~K. Bellamy, K.~Dey, M.~Hind, S.~C. Hoffman, S.~Houde, K.~Kannan, P.~Lohia, J.~Martino, S.~Mehta, A.~Mojsilovic, et~al.
\newblock Ai fairness 360: An extensible toolkit for detecting, understanding, and mitigating unwanted algorithmic bias.
\newblock \emph{arXiv}, 1810.01943, 2018.

\bibitem[Calmon et~al.(2017)Calmon, Wei, Ramamurthy, and Varshney]{calmon2017optimized}
F.~P. Calmon, D.~Wei, K.~N. Ramamurthy, and K.~R. Varshney.
\newblock Optimized data pre-processing for discrimination prevention.
\newblock \emph{arXiv}, 1704.03354, 2017.

\bibitem[Chen et~al.(2019)Chen, Kallus, Mao, Svacha, and Udell]{chen2019fairness}
J.~Chen, N.~Kallus, X.~Mao, G.~Svacha, and M.~Udell.
\newblock Fairness under unawareness: Assessing disparity when protected class is unobserved.
\newblock In \emph{Proceedings of the Conference on Fairness, Accountability, and Transparency}, pages 339--348, 2019.

\bibitem[COM(2021)]{com2021laying}
E.~COM.
\newblock Laying down harmonised rules on artificial intelligence (artificial intelligence act) and amending certain union legislative acts.
\newblock \emph{Proposal for a regulation of the European parliament and of the council}, 2021.

\bibitem[De~Larrard(2016)]{deLarrard}
A.~De~Larrard.
\newblock Commercial price optimization strategies in car insurance.
\newblock \emph{Insitut des actuaires}, 36\penalty0 (2):\penalty0 614--645, 2016.

\bibitem[Dwork et~al.(2011)Dwork, Hardt, Pitassi, Reingold, and Zemel]{Dwork2011}
C.~Dwork, M.~Hardt, T.~Pitassi, O.~Reingold, and R.~Zemel.
\newblock {Fairness Through Awareness}.
\newblock \emph{arXiv}, 1104.3913, 2011.
\newblock ISSN 0039-6109.
\newblock \doi{10.1145/2090236.2090255}.
\newblock URL \url{http://arxiv.org/abs/1104.3913}.

\bibitem[Grari et~al.(2021{\natexlab{a}})Grari, Hajouji, Lamprier, and Detyniecki]{grari2021learning}
V.~Grari, O.~E. Hajouji, S.~Lamprier, and M.~Detyniecki.
\newblock Learning unbiased representations via r{\'e}nyi minimization.
\newblock In \emph{Machine Learning and Knowledge Discovery in Databases. Research Track: European Conference, ECML PKDD 2021, Bilbao, Spain, September 13--17, 2021, Proceedings, Part II 21}, pages 749--764. Springer, 2021{\natexlab{a}}.

\bibitem[Grari et~al.(2021{\natexlab{b}})Grari, Lamprier, and Detyniecki]{grari2019fairness}
V.~Grari, S.~Lamprier, and M.~Detyniecki.
\newblock Fairness-aware neural r{\'e}nyi minimization for continuous features.
\newblock In \emph{Proceedings of the Twenty-Ninth International Conference on International Joint Conferences on Artificial Intelligence}, pages 2262--2268, 2021{\natexlab{b}}.

\bibitem[Grari et~al.(2022)Grari, Charpentier, and Detyniecki]{grari2022fair}
V.~Grari, A.~Charpentier, and M.~Detyniecki.
\newblock A fair pricing model via adversarial learning.
\newblock \emph{arXiv preprint arXiv:2202.12008}, 2022.

\bibitem[Hardt et~al.(2016)Hardt, Price, and Srebro]{hardt2016equality}
M.~Hardt, E.~Price, and N.~Srebro.
\newblock Equality of opportunity in supervised learning.
\newblock In \emph{Advances in neural information processing systems}, pages 3315--3323, 2016.

\bibitem[Hashorva et~al.(2018)Hashorva, Ratovomirija, Tamraz, and Bai]{hashorva2018some}
E.~Hashorva, G.~Ratovomirija, M.~Tamraz, and Y.~Bai.
\newblock Some mathematical aspects of price optimisation.
\newblock \emph{Scandinavian Actuarial Journal}, 2018\penalty0 (5):\penalty0 379--403, 2018.

\bibitem[Hu et~al.(2023)Hu, Ratz, and Charpentier]{hu2023fairness}
F.~Hu, P.~Ratz, and A.~Charpentier.
\newblock Fairness in multi-task learning via wasserstein barycenters.
\newblock In \emph{Joint European Conference on Machine Learning and Knowledge Discovery in Databases}, pages 295--312. Springer, 2023.

\bibitem[Ito and Fujimaki(2017)]{ito2017optimization}
S.~Ito and R.~Fujimaki.
\newblock Optimization beyond prediction: Prescriptive price optimization.
\newblock In \emph{Proceedings of the 23rd ACM SIGKDD international conference on knowledge discovery and data mining}, pages 1833--1841, 2017.

\bibitem[Kamiran and Calders(2012)]{kamiran2012data}
F.~Kamiran and T.~Calders.
\newblock Data preprocessing techniques for classification without discrimination.
\newblock \emph{Knowledge and Information Systems}, 33\penalty0 (1):\penalty0 1--33, 2012.

\bibitem[Lindholm et~al.(2022{\natexlab{a}})Lindholm, Richman, Tsanakas, and W{\"u}thrich]{lindholm2022discrimination}
M.~Lindholm, R.~Richman, A.~Tsanakas, and M.~V. W{\"u}thrich.
\newblock Discrimination-free insurance pricing.
\newblock \emph{ASTIN Bulletin: The Journal of the IAA}, 52\penalty0 (1):\penalty0 55--89, 2022{\natexlab{a}}.

\bibitem[Lindholm et~al.(2022{\natexlab{b}})Lindholm, Richman, Tsanakas, and Wüthrich]{lindholm2022discussion}
M.~Lindholm, R.~Richman, A.~Tsanakas, and M.~V. Wüthrich.
\newblock A discussion of discrimination and fairness in insurance pricing, 2022{\natexlab{b}}.

\bibitem[Lindholm et~al.(2023)Lindholm, Richman, Tsanakas, and Wuthrich]{lindholm2023fair}
M.~Lindholm, R.~Richman, A.~Tsanakas, and M.~V. Wuthrich.
\newblock What is fair? proxy discrimination vs. demographic disparities in insurance pricing.
\newblock \emph{Proxy Discrimination vs. Demographic Disparities in Insurance Pricing (May 2, 2023)}, 2023.

\bibitem[Lopez-Paz et~al.(2013)Lopez-Paz, Hennig, and Sch{\"o}lkopf]{lopez2013randomized}
D.~Lopez-Paz, P.~Hennig, and B.~Sch{\"o}lkopf.
\newblock The randomized dependence coefficient.
\newblock In \emph{Advances in neural information processing systems}, pages 1--9, 2013.

\bibitem[Louppe et~al.(2017)Louppe, Kagan, and Cranmer]{louppe2017learning}
G.~Louppe, M.~Kagan, and K.~Cranmer.
\newblock Learning to pivot with adversarial networks.
\newblock In \emph{Advances in neural information processing systems}, pages 981--990, 2017.

\bibitem[Moriah et~al.(2023)Moriah, Vermet, and Charpentier]{moriah2023measuring}
M.~Moriah, F.~Vermet, and A.~Charpentier.
\newblock Measuring and mitigating biases in motor insurance pricing.
\newblock \emph{arXiv preprint arXiv:2311.11900}, 2023.

\bibitem[Parliament and of~the European~Union(2016)]{EuropeanParliament2016a}
E.~Parliament and C.~of~the European~Union.
\newblock Regulation (eu) 2016/679 of the european parliament and of the council, 2016.
\newblock URL \url{https://data.europa.eu/eli/reg/2016/679/oj}.

\bibitem[Saxena et~al.(2024)Saxena, Zhang, and Shahabi]{saxena2024spatial}
N.~A. Saxena, W.~Zhang, and C.~Shahabi.
\newblock Spatial fairness: The case for its importance, limitations of existing work, and guidelines for future research.
\newblock \emph{arXiv preprint arXiv:2403.14040}, 2024.

\bibitem[Schmeiser et~al.(2014)Schmeiser, Störmer, and Wagner]{Schmeiser2014Unisex}
H.~Schmeiser, T.~Störmer, and J.~Wagner.
\newblock Unisex insurance pricing: Consumers’ perception and market implications.
\newblock \emph{The Geneva Papers on Risk and Insurance - Issues and Practice}, 39\penalty0 (2):\penalty0 322--350, 2014.

\bibitem[Treetanthiploet et~al.(2023)Treetanthiploet, Zhang, Szpruch, Bowers-Barnard, Ridley, Hickey, and Pearce]{treetanthiploet2023insurance}
T.~Treetanthiploet, Y.~Zhang, L.~Szpruch, I.~Bowers-Barnard, H.~Ridley, J.~Hickey, and C.~Pearce.
\newblock Insurance pricing on price comparison websites via reinforcement learning.
\newblock \emph{arXiv preprint arXiv:2308.06935}, 2023.

\bibitem[Verschuren(2022)]{verschuren2022customer}
R.~M. Verschuren.
\newblock Customer price sensitivities in competitive insurance markets.
\newblock \emph{Expert Systems with Applications}, 202:\penalty0 117133, 2022.

\bibitem[Wadsworth et~al.(2018)Wadsworth, Vera, and Piech]{wadsworth2018achieving}
C.~Wadsworth, F.~Vera, and C.~Piech.
\newblock Achieving fairness through adversarial learning: an application to recidivism prediction.
\newblock \emph{arXiv:1807.00199}, 2018.

\bibitem[Xin and Huang(2023)]{xin2023antidiscrimination}
X.~Xin and F.~Huang.
\newblock Antidiscrimination insurance pricing: Regulations, fairness criteria, and models.
\newblock \emph{North American Actuarial Journal}, pages 1--35, 2023.

\bibitem[Zafar et~al.(2017)Zafar, Valera, Rogriguez, and Gummadi]{Zafar2017mechanisms}
M.~B. Zafar, I.~Valera, M.~G. Rogriguez, and K.~P. Gummadi.
\newblock {Fairness Constraints: Mechanisms for Fair Classification}.
\newblock In \emph{AISTATS'17}, pages 962--970, Fort Lauderdale, FL, USA, 20--22 Apr 2017.

\bibitem[Zhang et~al.(2018{\natexlab{a}})Zhang, Lemoine, and Mitchell]{zhang2018}
B.~H. Zhang, B.~Lemoine, and M.~Mitchell.
\newblock {Mitigating Unwanted Biases with Adversarial Learning}.
\newblock \emph{Association for the Advancement of Artificial Intelligence}, jan 2018{\natexlab{a}}.
\newblock ISSN 15477401.
\newblock \doi{10.1080/08827508.2012.738731}.

\bibitem[Zhang et~al.(2018{\natexlab{b}})Zhang, Lemoine, and Mitchell]{zhang2018mitigating}
B.~H. Zhang, B.~Lemoine, and M.~Mitchell.
\newblock Mitigating unwanted biases with adversarial learning.
\newblock In \emph{AAAI'18}, pages 335--340, 2018{\natexlab{b}}.

\end{thebibliography}

\end{document}